\documentclass[letterpaper, 10 pt, conference]{ieeeconf}  
\usepackage{amssymb}
\usepackage{graphicx}
\usepackage{epstopdf}
\usepackage{float}
\usepackage{comment}

\newtheorem{remarkEnv}{Remark}

\usepackage[bb=dsserif]{mathalpha}

\IEEEoverridecommandlockouts                              

\usepackage{color}
\usepackage{amsmath} 

\DeclareMathOperator*{\Argmin}{Argmin}

\title{\LARGE \bf
Online Traffic Density Estimation using Physics-Informed Neural Networks
}

\author{Dennis Wilkman$^{1}$ and Kateryna Morozovska$^{1}$ and Karl Henrik Johansson$^{1}$ and Matthieu Barreau$^{1}$
\thanks{*This work is supported by the Wallenberg AI, Autonomous Systems and Software Program (WASP) funded by the Knut and Alice Wallenberg Foundation.}
\thanks{$^{1}$Authors are with the Division of Decision and Control Systems, School of Electrical Engineering and Computer Science,
        KTH Royal Institute of Technology, 100-44, Sweden
        {\tt\small \{dwilkman, kmor, kalle, barreau\}@kth.se}}%
}

\begin{document}

\maketitle
\thispagestyle{empty}
\pagestyle{empty}

\begin{abstract}

Recent works on the application of Physics-Informed Neural Networks to traffic density estimation have shown to be promising for future developments due to their robustness to model errors and noisy data. In this paper, we introduce a methodology for online approximation of the traffic density using measurements from probe vehicles in two settings: one using the Greenshield model and the other considering a high-fidelity traffic simulation. The proposed method continuously estimates the real-time traffic density in space and performs model identification with each new set of measurements. The density estimate is updated in almost real-time using gradient descent and adaptive weights. In the case of full model knowledge, the resulting algorithm has similar performance to the classical open-loop one. However, in the case of model mismatch, the iterative solution behaves as a closed-loop observer and outperforms the baseline method. Similarly, in the high-fidelity setting, the proposed algorithm correctly reproduces the traffic characteristics.

\end{abstract}

\section{Introduction}

The transport system accounts for approximately 18\% of global energy consumption \cite{owid-co2-emissions-from-transport}. Optimizing traffic flow can be one of the solutions to meet the UN's Sustainable Development Goal 11 by promoting sustainable and efficient urban mobility. The possibility of using rapid developments in autonomous vehicles for traffic state control allows reducing emissions while keeping travel time low by resolving traffic jams and also stabilizing human-controlled vehicles \cite{vcivcic2018traffic,delle2019feedback}. 

Similar to many other control strategies, effective autonomous transport requires an accurate estimation of the traffic density. Traditional methods require the use of sensor-based monitoring fixed on the road or camera surveillance and are therefore dependent on the surrounding infrastructure, and suffer from high costs, limited scalability, and challenges in real-time adaptability. One promising approach independent of infrastructure is to use the measurements from sensors directly on some vehicles traversing the road. However, this new trend relies on an increasing share of cars with probing capabilities. To maximize the benefits of mobile sensors, traffic can be modeled more effectively as a continuous 'fluid' instead of individual vehicles \cite{ferrara2018freeway}. And traffic density can be estimated using the first-order Lighthill-Withham-Richards (LWR) model \cite{RichardShock, lighthill1955kinematic}, by using hyperbolic partial differential equations (PDEs) to describe traffic flow. For a better match with real measurements, higher-order models have been developed in the high-fidelity simulation software SUMO \cite{sumo}.

Physics-informed neural networks (PINNs) have emerged as a promising approach for solving merging model and data-based solutions. PINNs are trained to solve supervised learning tasks while adhering to the physical laws of the given system \cite{karniadakis2021physics}. They can efficiently model traffic flow by combining the LWR model \cite{di2023physics} and density measurements sampled from probe vehicles \cite{barreau2021physics,huang2023physics}. PINNs are also more effective for traffic state estimation than the traditional control-based approaches, such as Extended Kalman Filters \cite{wang2005real} or open-loop observers \cite{barreau2020dynamic} when handling high-fidelity, sparse, and noisy offline data. 

In control applications, density estimation not only has to be accurate and robust, but it should also be achieved in almost real-time. Therefore, this paper presents a framework for real-time density estimation by adapting PINNs for online training. The resulting algorithm showcases the benefits of PINNs' application to the traffic density problem, while presenting various considerations such as the required number of past data-points to store, training time, and its effects on the performance of the estimation. We also investigate its performance in case of model mismatch on a high-fidelity SUMO simulation.

This manuscript is structured as follows. Section~\ref{sec2} contains the background on macroscopic traffic flow modeling and PINNs as well as the problem statement. Section~\ref{sec3} presents a methodology for training and evaluation of the online density estimation models, and outlines the assumptions considered for model training. Section~\ref{sec4} presents the results of the model implementation in two settings: the Greenshield model and high-fidelity simulations. When appropriate, a comparison between the PINN model and an open-loop observer is made. Finally, Section~\ref{sec5} summarizes the conclusions and outlines potential future research directions.


\section{Background and Problem Statement} \label{sec2}
This section describes the macroscopic and microscopic models used for modeling the dynamics of the traffic flow, as well as gives an introduction to PINNs implementation for traffic density reconstruction.

\subsection{Macroscopic and microscopic modeling of traffic flow}

Traffic flow can be modeled using different approaches, either with a microscopic traffic model by considering behaviors and interactions between the individual vehicles or with a macroscopic traffic model by disregarding the individual behaviors of the vehicles and instead focusing on the large-scale traffic properties.

Traffic flow modeled using LWR can be expressed using the continuity equation on the domain $\Omega = [0, T] \times \mathbb{R}$:
\begin{equation}
\centering
     \label{eq:LWR}
    \partial_t \rho + \partial_x\left( \rho v \right) = 0,
\end{equation}
where $\partial_t$ and $\partial_x$ are the partial derivative with respect to $t$ and $x$, respectively. The density $\rho(t,x)$ is the 'fluid' density or the number of vehicles at location $x$ and time $t$, $\rho \in [0, \rho_{max}]$. By definition, the maximum density $\rho_{max}$ is equal to the inverse of the average length of a vehicle plus the average distance between vehicles during a traffic jam. The velocity $v(\rho)$ is the instantaneous microscopic speed given density $\rho$. 

The velocity is equal to the free flow velocity at $0$ density, $v(0) = v_f$. It is a decreasing function, such that at the maximum density $\rho_{max}$, it should be equal to zero $v(\rho_{max}) = 0$ \cite{barreau2021physics}. 
This means that $\rho v(\rho)$ is a concave function that ensures the existence of a solution \cite[Theorem~6.2.2]{Dafermos:1315649} to the system \eqref{eq:LWR}. A very simple model for the microscopic velocity is given as $v = v_f\left(1-\frac{\rho}{\rho_{max}}\right)$ \cite{greenshields1935study}. 

Since \eqref{eq:LWR} does not have a unique solution and it might be discontinuous, it is common to consider the viscous form of \eqref{eq:LWR} instead. 
The viscous equation includes the diffusion term $\gamma \partial_{xx} \rho$ where $0 < \gamma \ll 1$, this ensures the numerical stability and uniqueness of the solution, together with smoother properties \cite{barreau2021learning}. The resulting macroscopic model with normalized density $\rho \in [0,1]$ on the same domain $\Omega = [0, T] \times \mathbb{R}$ writes as
\begin{equation}
\centering
    \label{eq:viscousLWR}
    \left\{ \begin{array}{l}
        \partial_t\rho + v_f (1 - 2\rho) \partial_x\rho = \gamma \partial_{xx} \rho, \\
        \rho(0, \cdot) = \rho_0.
    \end{array} \right.
\end{equation}

From a microscopic perspective, the penetration rate defines the fraction of available individual vehicles on the road equipped with probing sensors. If vehicle $i$ is one of them, then its position $x_i$ follows the dynamics equation
\begin{equation}
\centering
    \label{probevelocity}
    \dot x_i(t) = v_i(t, x_i) = v(\rho(t, x_i(t))). 
\end{equation}

Since both the micro and macro formulations are using the same velocity function $v(\rho)$ \cite{barreau2021learning}, the coupled micro-macro traffic model can be used to connect the behaviors of individual probe vehicles to the large-scale property density of the traffic across the entire road:
\begin{equation}
    \label{Coupled}
    \left\{
        \begin{array}{l}
            \partial_t \rho(t,x) = - v_f (1 - 2\rho) \partial_x\rho + \gamma\partial_{xx}\rho, \\
            \dot x_i(t) = v(\rho(t, x_i(t)).
        \end{array}
    \right.
\end{equation}

\begin{figure}[ht]
    \centering\includegraphics[width=0.95\linewidth]{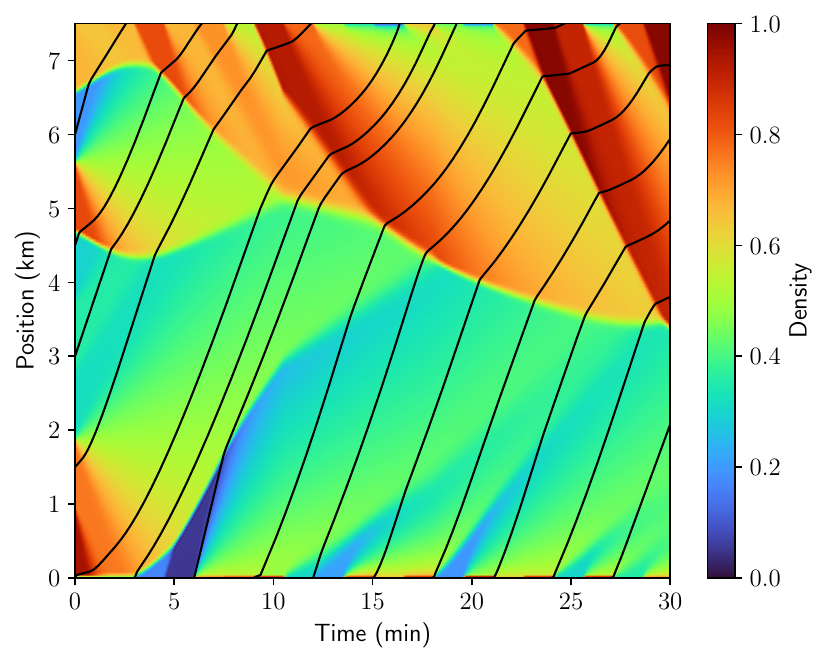}
    \caption{The black lines are the paths $x_i$ of the probe vehicles $i$, in which sensor data $\rho(\cdot, x_i(\cdot))$ is measured.}
    \label{fig:probe_demo}
\end{figure}

In Fig. \ref{fig:probe_demo} we see the simulated solution and the paths of the probe vehicles through space and time with the Greenshield closure equation $v(\rho) = v_f(1-\rho)$.

\subsection{Physics Informed Neural Network}
A feedforward neural network is defined as
\begin{equation}
    \centering
    \label{eq:nn}
    \mathcal{N}[\theta](\nu) = W_L H_{L-1} \circ H_{L-2} \circ \cdots H_1(\nu) + b_L,
\end{equation}
where $\theta = \{ (W_i, b_i) \}_{i=1, \dots, L}$ is the tensor of parameters, $W_i \in \mathbb{R}^{N \times N}$ is the weight, $b_i \in \mathbb{R}^{N \times 1}$ is the bias, $N$ is the number of neurons, $H_i(\nu) = \phi(W_i \nu + b_i)$ is the layer $i$, $\phi$ is an element-wise activation function and $L$ is the number of layers. The first and last layers have appropriate dimensions according to the input and output dimensions of the network. The universal approximation theorem \cite{hornik1991approximation} states that a feedforward neural network with $L > 0$ can approximate arbitrarily close any smooth function, provided a sufficient number of neurons $N$.

The PINN is trained to approximate a solution and/or parameters of the PDE from the data. PINN is defined as the optimization under constraints problem
\begin{align}\label{eq:optimization_PINN}
    \theta^* = \begin{array}[t]{cl} \Argmin_{\theta} 
        & \displaystyle \frac{1}{| \mathcal{D} |} \sum_{(\nu, y) \in \mathcal{D}} \| y - \mathcal{N}[\theta](\nu) \|^2 \\
        \text{s.t.} & \displaystyle \int_{\Omega} \| \mathfrak{F}\left[\mathcal{N}[\theta]\right](\nu)\|^2  d\nu = 0,
    \end{array}
\end{align}
where the dynamical constraints, expressed as differential equations, are defined as $\mathfrak{F}[\mathcal{N}[\theta]] = 0$ on $\Omega$. The measurements are formulated as a dataset $\mathcal{D} = \left\{ (\nu_i, y_i) \right\}_{i}$ where $\nu_i$ is the input giving the output $y_i$.

Problem~\eqref{eq:optimization_PINN}, however, is not numerically tractable. A solution is to relax it, leading to:
\begin{multline}\label{relax}
    \min_{\theta} \ \max_{\Lambda} \frac{1}{| \mathcal{D} |} \sum_{(\nu, y) \in \mathcal{D}} \| y - \mathcal{N}[\theta](\nu) \|^2 \\
        + \frac{1}{|\mathcal{D}_{\mathfrak{F}}|} \sum_{\nu \in \mathcal{D}_{\mathfrak{F}}} \left\|\mathfrak{F}\bigl[\mathcal{N}[\theta]\bigr](\nu) \right\|_{\Lambda}^2,
\end{multline}
where $\mathcal{D}_{\mathfrak{F}} \subset \Omega$ is a uniform sampling over $\Omega$ and $\|A\|_{\Lambda}^2 = A^{\top} \Lambda A$ for $\Lambda$ definite positive. In that formulation, $\Lambda$ is a positive diagonal matrix representing the Lagrange multipliers and is introduced to penalize divergence of the physics residuals. Problem~\eqref{relax} is a learning problem that can be efficiently solved using gradient-descent on $\theta$ and gradient-ascent on $\Lambda$ \cite{barreau2025accuracy}.

\subsection{PINNs and traffic state reconstruction}
Based on the original work in \cite{barreau2021physics}, the PINN model for traffic density estimation can be defined as \eqref{eq:optimization_PINN} with
\begin{align}
    \label{pi}
    \hat{\rho}[\theta](t, x) &= \mathcal{N}[\theta]\left( \left[\begin{matrix} t,  x \end{matrix}\right] \right), \\
    \hat{v}[\psi](\rho) &= (1 - \rho) \left( v_f + \rho \mathcal{N}[\psi](\rho)^2 \right), \\
    \mathfrak{F}[\hat{\rho},\hat{v}] &= \left[ \begin{matrix} \partial_t\hat{\rho} + \left( \hat{v} + \rho \hat{v}' \right) \partial_x\hat{\rho} - \gamma \partial_{xx} \hat{\rho} \\ \max(\hat{v}', 0) \end{matrix} \right],
\end{align}
with $\hat{v}' = \frac{d\hat{v}}{d\rho}$ and $(t, x) \in \Omega = [0, T] \times [0, L]$. 

\begin{figure}
    \centering
    \includegraphics[width=1\linewidth]{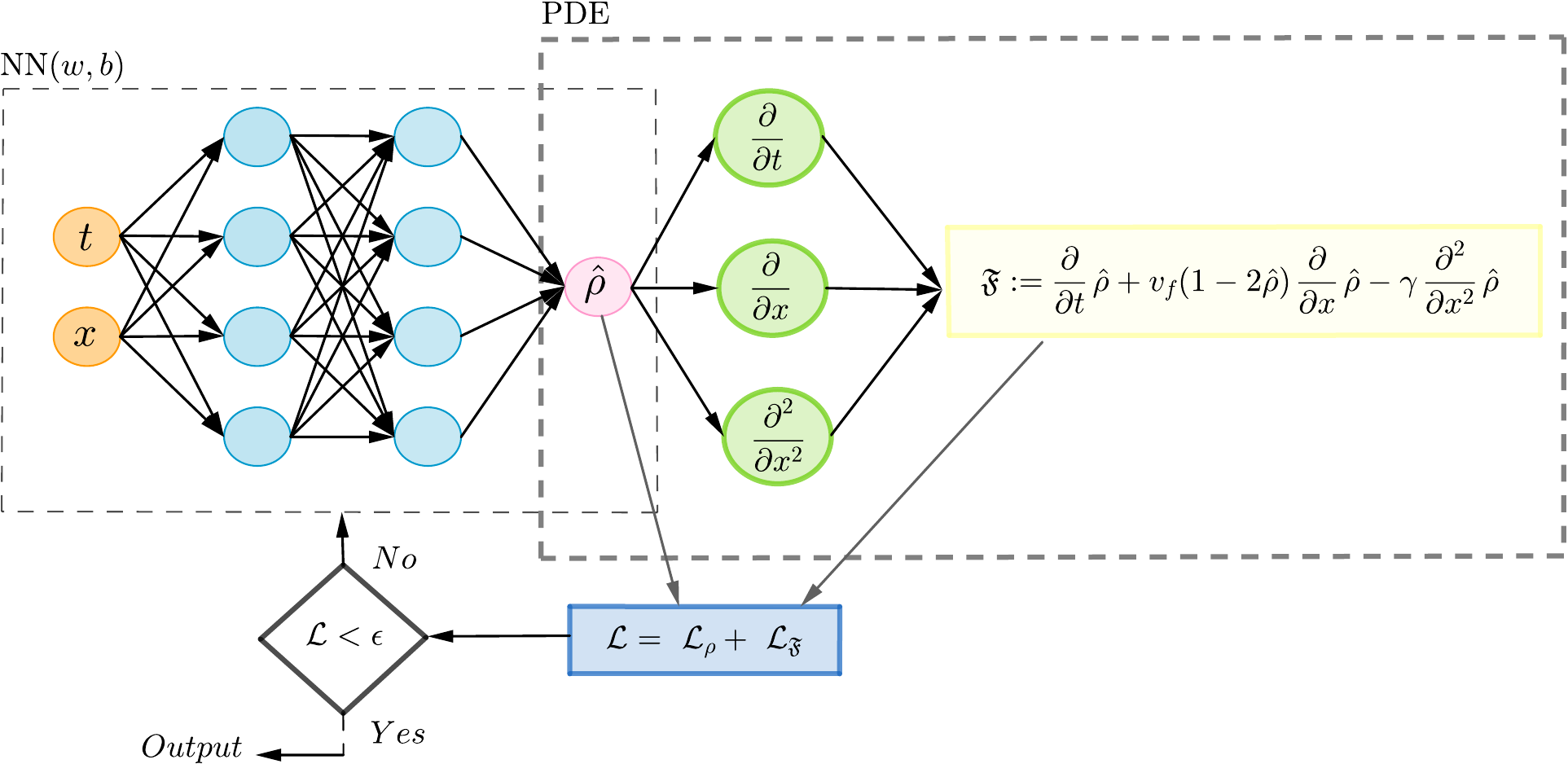}
    \caption{The PINN for traffic density reconstruction.}
    \label{fig:PINN}
\end{figure}

The dataset $\mathcal{D} = \mathcal{D}(0, T)$ is defined as:
\begin{equation}
    \mathcal{D}(t_1, t_2) = \underset{i \in \{1, \dots, N\}}{\bigcup} \mathcal{D}_i(t_1, t_2),
\end{equation}
where 
\begin{equation}
 \!\!\!   \mathcal{D}_i(t_1, t_2) = \{ (t_k, x_i(t_k), v_i(t_k), \rho(t_k, x_i(t_k) \}_{t_k \in [t_1, t_2]},
\end{equation}
where $\{t_k\}$ are the sampling times and $v_i(t_k)$ is the speed of probe vehicle $i$ at time $t_k$. The set $\mathcal{D}_i(t_1, t_2)$ refers to the data collected by probe vehicle $i$ between time $t_1$ and $t_2$. The data loss function can then be expressed as 
\begin{multline}
    \mathcal{L}_{\rho}(t_1, t_2) = \sum_{i = 1}^N \sum_{(t, x, v, \rho) \in \mathcal{D}_i(t_1, t_2)} \left\{ \bigl| \hat{\rho}[\theta](t, x) - \rho \bigr|^2 \right. \\
    \left.+ \bigl| \hat{v}[\psi](\rho) - v \bigr|^2 \right\},
    \label{lossnew}
\end{multline}
while the physics loss function is defined as:
\begin{equation}
    \mathcal{L}_{\phi}(t_1,t_2,\Lambda) = \frac{1}{|\mathcal{D}_{\mathfrak{F}}(t_1, t_2)|} \!\!\sum_{\nu \in \mathcal{D}_{\mathfrak{F}}(t_1, t_2)} \!\!\!\!\!\!\! \left\|\mathfrak{F}\bigl[\mathcal{N}[\theta]\bigr](\nu) \right\|_{\Lambda}^2,
    \label{lossdata}
\end{equation}
where $\mathcal{D}_{\mathfrak{F}}(t_1, t_2)$ is a uniform sampling of collocation points over $[t_1, t_2] \times [0, L]$. In Fig. \ref{fig:PINN}, we illustrate the general training method in this setting, formulated by the following optimization problem:
\begin{equation}
    \label{eq:tse_pinn}
\!\!    \theta^*(t_1, t_2, \delta) \!\! = \!\! \Argmin_{\theta} \max_{\Lambda} \!\! \ \mathcal L_{\rho}(t_1, t_2) +  \mathcal{L}_{\phi}(t_1, t_2+\delta, \Lambda)
\end{equation}
and the optimal reconstruction is $\hat{\rho}[\theta^*(0, T, \delta)]$.\\
Note that this approach is particularly well suited to ill-posed problems where there are not enough or too many measurements to ensure the existence of a unique solution. In our case, probe vehicle measurements are introducing redundancy and then possibly over-constraining the problem. At the same time, this redundancy can be exploited to uncover some unknown parts of the model dynamics, making this framework adaptable to model identification \cite{barreau2021physics}. 

Another important aspect of this method is the extension to prediction. In that case, future values of the density are estimated using the model $\mathfrak{F}[\hat{\rho},\hat{v}] = 0$ over a larger domain $[0, T + \Delta T] \times [0, L]$. Therefore, the neural network 
\begin{equation}
    \label{eq:rhoT}
    \hat{\rho}_T = \hat{\rho}\bigl[ \theta^*(0, T, \Delta T) \bigr]
\end{equation}
can be meaningfully evaluated on $[0, T+\Delta T]$ even if no measurements are collected during $[T, T + \Delta T]$. 

\subsection{Problem Statement}
The traffic state reconstruction described in the previous subsection shows great estimation capabilities, with model identification and prediction naturally included in the method. However, approximating the solution using gradient descent is a very slow and computationally demanding approach. This leads to longer training times, which prevents its use for real-time problems.

The objective of the paper is to find a causal algorithm to continuously give an accurate estimate of the real-time traffic density in real-time. Given the scenario that new sensor measurement data is sampled in real-time from probe vehicles. We are interested in minimizing the Current Estimation Error ($CEE$) between the true density $\rho(t,\cdot)$ and its estimation $\hat\rho(t, \cdot)$, given as:
\begin{equation}
    \label{eq:CEE}
    CEE_t(\rho, \hat\rho) = \int_{0}^{L} |\rho(t, x) - \hat{\rho}(t, x)|^2dx.
\end{equation}

Note that causality implies that, at time $t$, the estimation $\hat{\rho}$ only uses the data available until the same time $t$.

Another objective is to keep the $CEE$ low even in case of model mismatch, a difference between the assumed and true dynamics of the system. The proposed method should be able to adapt in real-time to reduce its impact on the estimation error. 
\section{Methodology}\label{sec3}
In this section, we cover a general framework for real-time density estimation with PINNs, then we discuss some fundamental limitations and investigate how training can be adapted to that purpose. 
\subsection{Framework for online real-time estimation through iterative learning}
The main idea behind this framework is to leverage the predictive capabilities of PINNs as a buffer to perform offline operations. 

At time $T > 0$, we have data obtained from probing vehicles in the time window $[0, T]$. If we begin training the PINNs at time $T$ on the domain $[0, T+\Delta T]$, we get $\hat{\rho}_T$ after solving the optimization problem~\eqref{eq:tse_pinn}. Note that $\Delta{T}$ is a parameter to be chosen. Since training and inferring from the neural network is a slow process, it will take $\delta_t$ seconds and the estimate will be ready at time $T+\delta_t$.

Therefore, from time $T + \delta_t$ we will have the predictions trained on the data between time $[0, T]$. If we train a new model continuously, then at time $T + \delta_t$, we will begin training a new model for which inference will be available at time $T + 2 \delta_t$. Consequently, to have an estimate at any time between $[T, T + 2 \delta_t]$, the first training should include prediction over $[T, T + 2 \delta_t]$, meaning that $\Delta T = 2 \delta_t$. In that case, we can use equation~\eqref{eq:rhoT} and $\hat{\rho}_{\delta_t}$ will produce the estimate for $t \in [T + \delta_t, T + 2 \delta_t]$. Pursuing this logic and assuming that $T = \delta_t$, we will use the following reconstructed state for any $t$ larger than $2\delta_t$:
\begin{equation}
\centering
\label{rhohat}
    \hat{\rho}(t, \cdot) = \left\{ \begin{array}{ll}
        \hat{\rho}_{\delta_t}(t, \cdot) & \text{ if } t \in [2\delta_t, 3\delta_t), \\
        \quad \vdots\\
        \hat{\rho}_{i\delta_t}(t, \cdot) & \text{ if } t \in [(i+1)\delta_t, (i+2)\delta_t), \\
        \quad \vdots
    \end{array} \right.
\end{equation}

\begin{figure}[ht]
    \centering
    \includegraphics[width=1\linewidth]{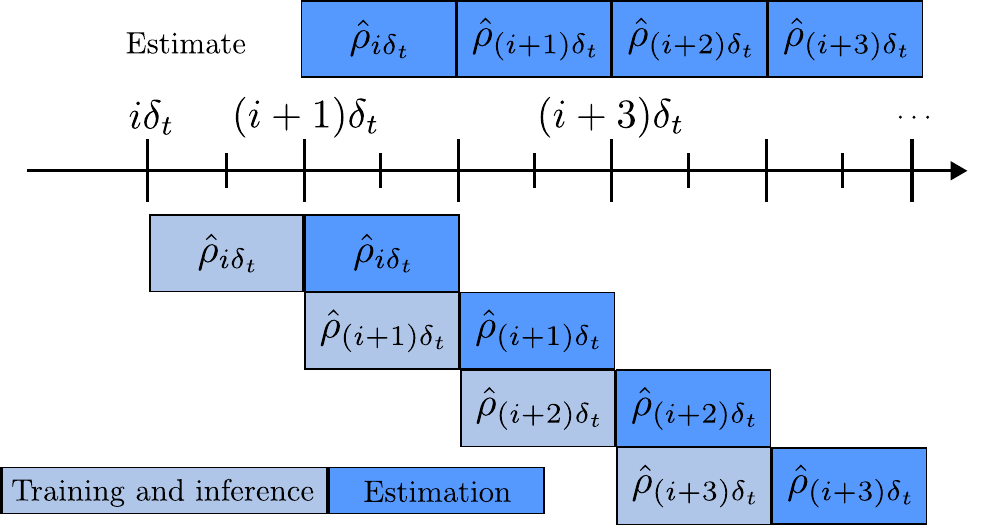}
    \caption{Framework for using a PINN for online real-time estimation using iterative learning.}
    \label{fig:framework}
\end{figure}

Fig. \ref{fig:framework} illustrates the process of continuously training models using the proposed framework.
\subsection{Practical considerations}
The previous subsection introduced a framework for using PINNs in a real-time setting. However, it also has certain limitations, which are addressed in this subsection.
\subsubsection{Infinite horizon training}
In the given framework, the size of the inference domain continues to increase with each iteration. At time $i \delta_t$ with $i > 2$, the size of the PINN domain is equal to $(i+2) \delta_t L$. As defined in \cite[Proposition~1]{barreau2025accuracy}, the accuracy of a neural network is, at worst, proportional to the size of the inference window. This implies a larger neural network to enable a low $L^2$ error over the domain, leading to the first research question.

\noindent \textbf{Question~1:} How to keep the size of the neural network constant with a constant accuracy?

\subsubsection{Training Time}

A major challenge that arises when estimating in real-time is that training time itself impacts the quality of predictions. Hence, if the model has completed training and inferred at time $t$, then it will only have been trained on data up to the point $t-\delta_t$. The longer the training time $\delta_t$, the less relevant the data in the training set becomes. Therefore, computationally heavy methods that traditionally are expected to improve the accuracy of the estimation may instead have the opposite effect due to the increased training time $\delta_t$. 


For the proposed framework, the increasing domain size will increase the number of training iterations (larger neural network) and the number of data points. Assume that the training time $\delta_t$ at time $T$ is approximately equal to $e_t(\alpha|\mathcal{D}_{\mathfrak{F}}(0, T+2\delta_t)| + \beta|\mathcal{D}(0,T)|)$, where $e_t$ is the number of training iterations, and $\alpha, \beta$ are determined by factors such as network architecture and training setup. Then, $\delta_t$ will be unbounded and, depending on the number of physics collocation points used, the effects on the training time from the number of data points in the data set $|\mathcal{D}_{\mathfrak{F}}(t_1, t_2)|$ can vary substantially. This leads to the following research question.

\noindent \textbf{Question~2:} How to ensure the accuracy of PINNs with constant training time?


%

\subsubsection{Robustness}

Traffic can, at times, be unpredictable. With the LWR framework, the free flow velocity $v_f$ might change with time and space. This would, in turn, reduce the quality of the estimation due to a mismatch between the real and assumed model. The same applies to sensor noise, which can be affected by the current conditions. This leads to the following question.

\noindent \textbf{Question~3:} How to keep the model robust against model errors in an online setting?

\subsection{Extending PINNs for online learning}

We introduce adaptations of the classical PINNs training to extend to practical online learning by proposing answers to the three questions raised previously.

\subsubsection{Training time-window}

The size of the dataset $\mathcal{D}$ grows with time, which eventually causes the training time $\delta_t$ to become unreasonably large, and the increasing size of the training domain causes the neural network to produce worse results. The simplest solution is to introduce a maximum look-back time $\delta_d$. This modification creates a moving-time window $[T-\delta_d, T]$ for which the past data is stored. Using formulation \eqref{eq:rhoT}, we introduce:
\begin{equation}
    \centering
    \label{rhoT}
    \hat{\rho}_{T, \delta_d} = \hat{\rho}\bigl[ \theta^*(T-\delta_d, T, 2\delta_t) \bigr].
\end{equation}
Therefore, if a model is continuously trained, the new reconstructed density is:
\begin{equation}
\centering
\label{reconstructed}
    \hat{\rho}(t, \cdot) = \hat{\rho}_{i \delta_t, \delta_d}(t, \cdot),
\end{equation}
where $i = \lfloor t/\delta_t \rfloor-1$.

This approach implies that at time $T$, we forget about data gathered before $T-\delta_d$. Based on \cite{delle2019traffic}, there always is a time window $\delta_d$ such that \eqref{Coupled} has a unique solution after time $T$ using measurements from probing vehicles as boundary conditions. 
However, $\delta_d$ can be arbitrarily large if the densities are small. As a consequence, the time window introduces a trade-off between the uncertainty in the reconstruction at low densities, the size of the neural network, and the training time.

\subsubsection{Efficient training of PINNs in an online setting}  
Here we consider using the time window defined above. At the beginning of the $i$th training, starting at time $i\delta_t$, consider the time-window of the data to be $\mathcal{I}_i = [i\delta_t-\delta_d, i\delta_t]$ with $\delta_d>\delta_t$, to ensure no times exist where no data is collected. The optimization~\eqref{eq:tse_pinn} is then conducted from scratch, using, for instance, Xavier initialization for the parameters $\theta$. This is a waste of computational resources since the model trained at time $(i-1)\delta_t$ considered the time-window $\mathcal{I}_{i-1}$ and $\mathcal{I}_{i-1} \cap \mathcal{I}_{i}$ is not empty. Consequently, the previous model contains valuable information that can be transferred to the new model. This could serve both as a warm startup, which will significantly decrease the training time, and act as propagation of the initial condition from the last time step. 

Reusing a previous model for the new training is possible by doing a normalization and then a shift. First, since the time input for a given model $i$ belongs to the window $[i\delta_t-\delta_d, (i+2)\delta_t]$, one can define the transformation:
\begin{equation}
\label{si}
\centering
     s^{(i)}: t \mapsto \frac{2}{2 \delta_t + \delta_d}\left(t - (i+1)\delta_t + \frac{\delta_d}{2} \right),
\end{equation}
which will ensure that $s^{(i)}(t) \in [-1, 1]$ if $t \in [i\delta_t-\delta_d, (i+2)\delta_t]$. This normalization acts as a pre-processing step to guarantee that the input is bounded. Then, it is possible to keep almost all the parameters from the previous training up to a change in the bias $b_1$ to introduce a time shift:
\begin{equation}
    \centering
    \label{timeshift}
    \hat{\rho}_{i\delta_t,\delta_d}\left( (i+2)\delta_t, \cdot \right) = \hat{\rho}_{(i+1)\delta_t,\delta_d}\left( (i+2)\delta_t, \cdot \right).
\end{equation}
If the bias of the first layer for the iteration $i$ is denoted by $b_1^{(i)}$ given $\mathbbb{1} = \left[ \begin{matrix}
    1 & 0
\end{matrix} \right]$, then the previous equality implies:
\[
    \mathbbb{1} W_1 s^{(i)}((i+2)\delta_t) + \mathbbb{1} b_1^{(i)} = \mathbbb{1} W_1  s^{(i+1)}((i+2)\delta_t) + \mathbbb{1} b_1^{(i+1)},
\]
for any $x \in [0, L]$, which leads to:
\[
    b_1^{(i+1)} = b_1^{(i)} + \left[
    \begin{matrix} \mathbbb{1} W_1 \left[ s^{(i)}((i+2)\delta_t) - s^{(i+1)}((i+2)\delta_t) \right] \\ 0 \end{matrix} \right].
\]
By using this transformation before the next iteration of training, the information from before $i\delta_t-\delta_d$ is also preserved as a warm startup, allowing for a potentially greater information retention of data before the window size $\delta_d$.

\begin{figure}[ht]
    \centering
    \includegraphics[width=0.8\linewidth]{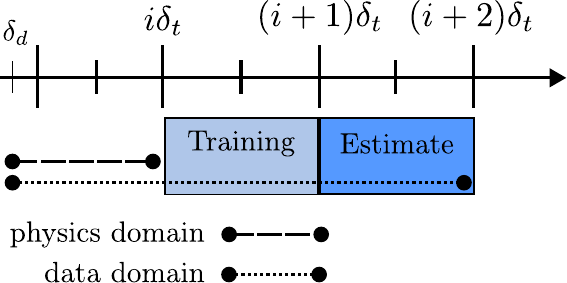}
    \caption{The time scale for training a new PINN iteration at time $i\delta_t$.}
    \label{fig:enter-label}
\end{figure}

\subsubsection{Model identification capabilities in an online setting}

In the online case, it is reasonable to assume that  $v_f$ can vary with time, due to a variety of factors such as road accidents or weather conditions. Since an estimate of the free flow velocity is required for our framework, we need to learn it in real-time. However, it is not physically accepted that $v_f$ varies at each training iteration. Therefore, we propose to consider the window size $\delta_v$ for fitting velocity measurements based on two considerations: a short window allows for faster adaptation, while a longer one increases the size of the data set. The second option makes it more robust to noisy measurements of traffic densities and velocities because it contains a wider range of densities, which also allows the PINN to learn more complex traffic models.

\section{Results}\label{sec4}
The results of this paper consist of the detailed description of the model implementation and the results of the numerical experiments using Greenshield model as well as on data from SUMO simulations that better mimic realistic traffic scenarios. The proposed PINN framework applied to the Greenshield model is compared with a traditional open-loop observer for different degrees of model mismatch, and we investigate the impact of training time on the error. Later, the results also discuss model's performance with SUMO. 
\subsection{Greenshield model}
\subsection*{\textbf{Model implementation:}}
The time windows of the data and velocity measurements are set to $\delta_d = \delta_v = 3 \space min$, the sampling rate $f_s$ is set to 3 samples per second, the number of training iterations $e_t$ is set to 100 epochs, and the neural network has $L = 2$ layers and $N = 32$ neurons making 
$\delta_t = 0.3 \space min$. Note that $\delta_t$ is dependent on the available computing power. 

The simulation data is generated by solving \eqref{eq:viscousLWR} using a finite difference method \cite{NumericalConservative} with a maximum value of $v_f=37.5\space km/h$, and the boundary conditions of $\rho$ as piecewise constants in the range $[0, 1]$. The free flow velocity varies with time such that:
\[
    v_f(t) = \left\{ \begin{array}{ll}
    37.5 \space km/h & \text{ if } t \in [0, 10), \\
    18.75 \space km/h & \text{ if } t \in [10, 18), \\
    30 \space km/h & \text{ if } t \in [18, 30]. \\
    \end{array} \right.
\]

For the real-time PINN, the free flow velocity $v_f$ is learned from data, and the updates are sufficiently quick so that $\delta_t$ is assumed to be equal to the sampling rate. We compare our results with the open-loop observer designed in \cite{barreau2020dynamic} with specified $v_f = 37.5\space km/h$. 


\subsection*{\textbf{Numerical experiments:}}

\begin{figure*}[!t] 
    \centering
    \includegraphics[width=0.8\linewidth]{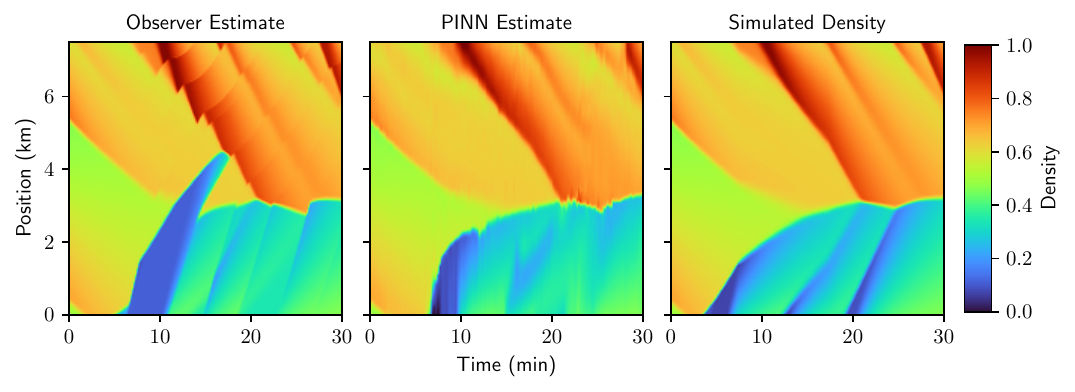}
    \caption{Reconstructed and simulated data, in the case of varying free flow velocity, leading to areas of imperfect model knowledge.}
    \label{fig:data}
\end{figure*} 
\subsubsection{Comparison between open-loop observer and PINN with varying mismatches in model knowledge}

\begin{figure}[ht]
    \centering
    \includegraphics[width=0.95\linewidth]{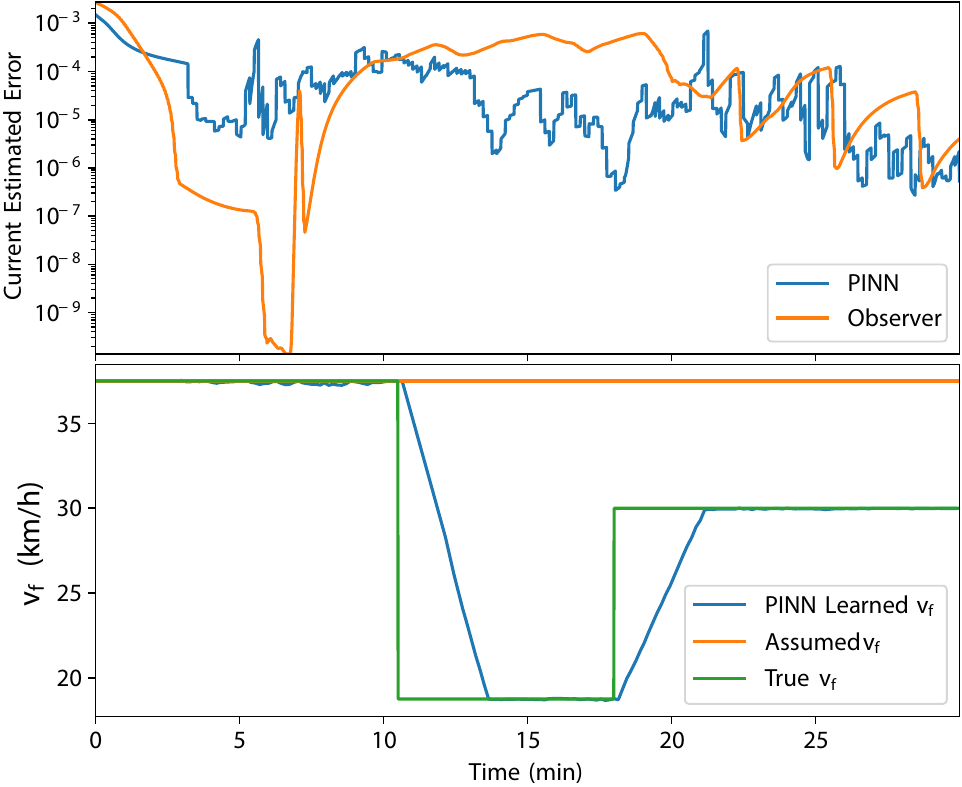}
    \caption{$CEE$ for the observer and PINN for a varying  $v_f$.}
    \label{fig:error-error}
\end{figure}

Fig. \ref{fig:data} shows the results with a varying free flow velocity, for the observer and the respective real-time estimates of the observer and the PINN. Fig. \ref{fig:error-error} illustrates the resulting $CEE$ across time for the observer and PINN. In the initial time between $t \in [0, 7]$, when the traffic density is higher, we can see how the observer error is decreasing rapidly, while the PINN error fluctuates, most likely due to the lack of training iterations required for convergence. Between $t \in [7, 10]$ we can see the error is spiking and increasing for the observer as the boundary condition changes to low-density flow. In the interval of the perfect model knowledge $t \in [0,10]$, the observer outperforms the PINN model. The lower density also affects the PINN error since the solution is not unique given the probe vehicle data. Moreover, PINN has a minimum CCE given a neural architecture because of its limited expressivity.\\
Starting with the new interval $t \in [10, 18]$, we notice the artifacts in the observer estimation in case of the model mismatch, and the resulting error increases. For the PINN model, the error stagnates until the learned free flow velocity converges to the new value, and afterwards the error begins to rapidly decrease. A contributing factor to the reduced error can also be explained by the increase in density of the lower boundary conditions. In this time frame, in the case of a large model mismatch, we can see that PINN achieves a strictly lower error by adapting to the changing free-flow velocity.

In the last interval $t \in [18, 30]$, the error starts to increase for the PINN model since the learned free flow velocity $v_f$ has an initial high discrepancy, while the error decreases for the observer due to the reduced model discrepancy. From $t \in [21,30]$, both the observer and the PINN model errors decrease gradually with time, achieving comparable performance. At $t = 21$min the PINN model no longer has any velocity or density measurements from the previous free flow velocity in the data window, which explains why the error starts to decrease. A lower $\delta_d$ or $\delta_v$ would have contributed to a faster recovery in error after the free flow shift at the price of a lower robustness. From the following simulations, we can conclude that in the case of perfect model knowledge, the observer has superior performance; however, in the case of high model error, the PINN model outperforms the observer, and with a low model error, the performances are similar.


\subsubsection{The effect of parameter changes on performance}
\begin{figure}[h]
    \centering
    \includegraphics[width=0.95\linewidth]{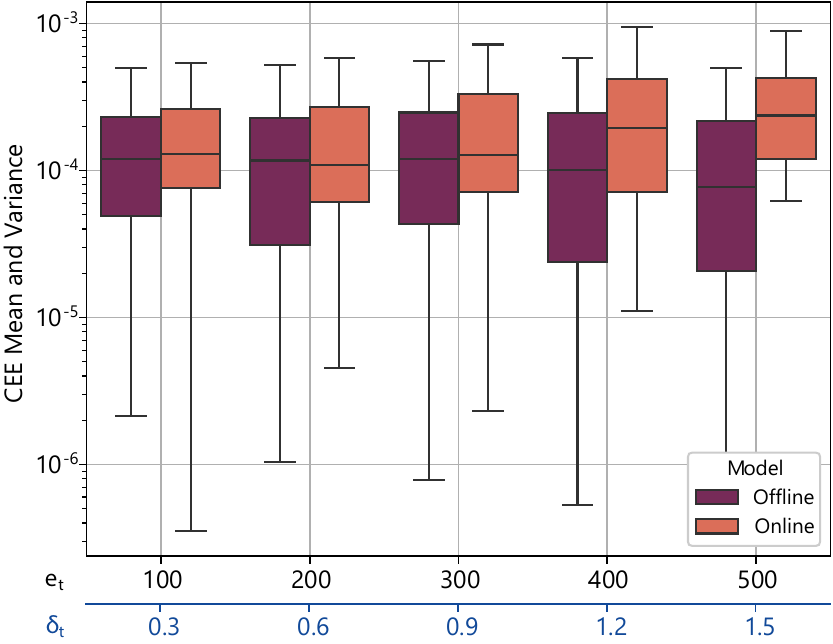}
    \caption{$CEE$ for different training iterations $e_t$, one model trained in the `Online' case, such that $\delta_t$ increases with $e_t$, and another model trained in the `Offline' case, such that $\delta_t=0.3$ independent of $e_t$.}
    \label{fig:iter}
\end{figure}

Fig. \ref{fig:iter} presents the comparison of results after training two models, an `Online' model, in which $\delta_t$ depends on the number of training iterations, and an `Offline' model, where $\delta_t=0.3$ is kept constant. Here we can see that the mean $CEE$ error for both 'Online' and `Offline' models are similar, when $e_t\in[100,300]$. However, when $e_t \geq 300$,  the `Online' model error starts to increase, and the `Offline' model error starts to decrease. Hence Fig. \ref{fig:iter} highlights the trade-off appearing in the online case, where the increased number of iterations leads to increased error, contrary to the traditional scenario, when the error is expected to decrease. 
\begin{figure*}[!t] 
    \centering
    \includegraphics[width=0.95\linewidth]{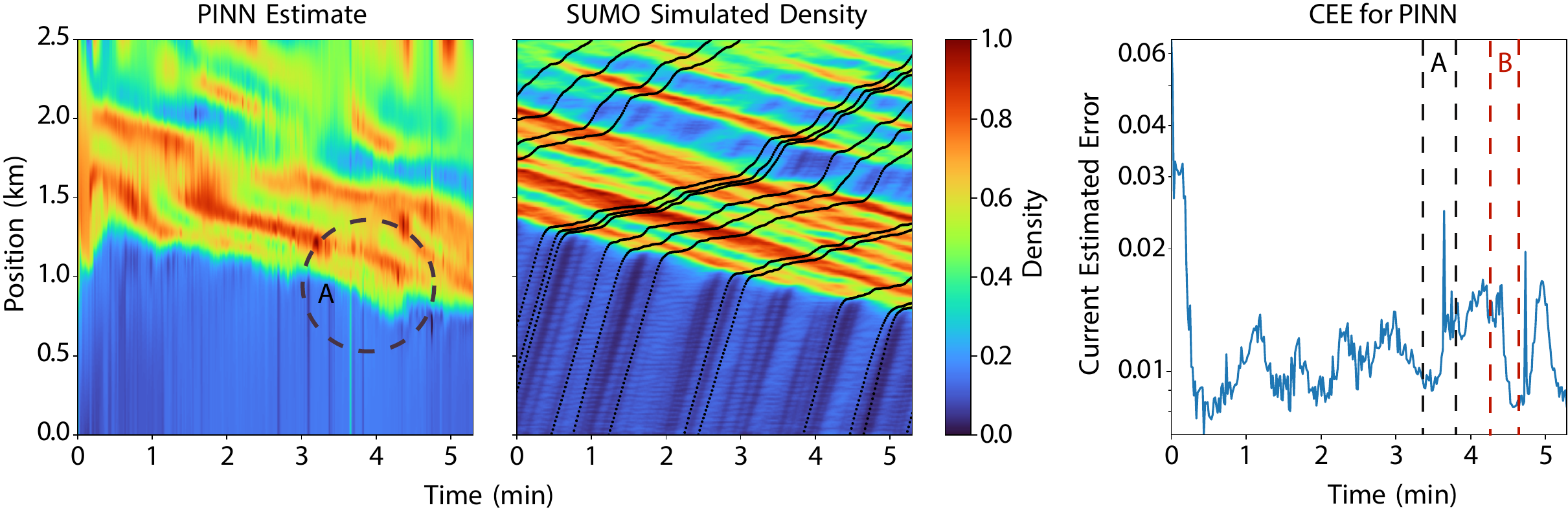}
    \caption{The estimation of the current time density is shown on the left. The corresponding SUMO simulation with dots marking the measurements from the probe vehicles is depicted in the central figure. On the right we see the corresponding $CEE$ between the PINN estimates and SUMO simulation.}
    \label{fig:sumo-sim}
\end{figure*}

\subsection{SUMO simulation}
SUMO simulation allows testing our model for robustness in a more realistic scenario. The exact dynamics of the macroscopic behavior of the system is unknown. As such the velocity function $v(\rho)$ has to be learned, adhering to the constraints $v(0)=v_f$, $v(1) = 0$ and $\partial_\rho v<0$.

The PINN estimate of the current density, the true density produced by a SUMO simulation and the $CEE$ of the PINN can be seen in Fig. \ref{fig:sumo-sim}. 
The PINN reaches initial convergence at around $t=0.5 \space min$ after which the $CEE$ oscillates around the point $CEE\approx0.01$. A possible explanation to the oscillations is due to discontinuities in the data, for example, in the area marked as 'A' in Fig. \ref{fig:sumo-sim}, we can see the corresponding error increase, and then a subsequent decrease in the area marked as 'B'. Area 'A' does not have any probe vehicle data and, as time progresses, we can see how an erroneous green 'tail', is formed, due to previous errors being propagated. As the tail is forming we can see the error increase with time until approximately $t\approx 4.4 min$, where a probe vehicle arrives to supply new measurements. Subsequently, the tail disappears, and the error rapidly declines. 

Another behavior that is notable, is the short spike in error appearing at time $t\approx 3.6 min$ and $t\approx4.7$. From the figure we can see the error at $t\approx 3.6$ caused by the road segment $x\in[0, 1.0]$ having an incorrectly higher density compared to its neighborhood. 
This potentially could be caused by either a lack of data from probing vehicles or insufficient training iterations not allowing the information to propagate across the area or a combination of the two. 

\section{Conclusion}\label{sec5}
The demand for accurate real-time traffic density estimations is rising due to the need for effective autonomous transport and the increased reliance on the implementation of traffic control strategies.

To address the issues arising from moving to real-time traffic control, we propose a framework for online traffic density estimation using PINNs. The conducted numerical experiments highlight the robustness of the PINN in the case of a time-varying traffic model, when the PINN outperforms a classical open-loop observer. They also show the negative effects of training time on the estimation quality, a property only apparent in the online scenario. The application to data generated using SUMO shows that the proposed method performs well, even in the case of an unknown high-fidelity dynamical model.

Future research on the topic should focus on improving the speed of convergence of the PINN to reduce the impact of the training time, investigate the introduction of uncertainty quantification identify areas of high and low reliability of the estimate, and improving the adaptability of the network by considering the use of dynamic weights on the probe vehicle measurements.

\addtolength{\textheight}{-12cm}   








\bibliographystyle{ieeetr}
\bibliography{reference}

\end{document}